\documentclass[
    conference,
    10pt,
    ]{IEEEtran}
\usepackage[utf8]{inputenc}
\usepackage[T1]{fontenc}

\usepackage[ruled,vlined,linesnumbered]{algorithm2e}
\usepackage{amsmath}
\DeclareMathOperator*{\argmax}{arg\,max}

\usepackage{amsthm}
\usepackage{flushend}
\usepackage{graphicx}
\makeatletter
\@namedef{ver@lineno.sty}{9999/12/31}
\@namedef{opt@lineno.sty}{}
\makeatother
\usepackage[draft,outputdir=build,finalizecache]{minted}  
\usepackage{siunitx}
\usepackage{xcolor}

\makeatletter
\let\MYcaption\@makecaption
\makeatother
\usepackage[font=footnotesize]{subcaption}
\makeatletter
\let\@makecaption\MYcaption
\makeatother

\usepackage{hyperref}
\usepackage[capitalise]{cleveref}

\newcommand{\merge}[1]{{#1}}
\title{RL-GRIT: Reinforcement Learning \\ for Grammar Inference}
\author{%
\IEEEauthorblockN{Walt Woods}%
    \IEEEauthorblockA{\small Galois, Inc.\\
        \texttt{waltw@galois.com}}
}
\date{February 2021}

\let\includegraphicsold\includegraphics
\renewcommand{\includegraphics}[1]{\includegraphicsold[width=\linewidth]{#1}}

\begin{document}

\date{}
\maketitle
\thispagestyle{plain}
\pagestyle{plain}

\begin{abstract} 
    When working to understand usage of a data format, examples of the data format are often more representative than the format's specification. For example, two different applications might use very different JSON representations, or two PDF-writing applications might make use of very different areas of the PDF specification to realize the same rendered content. 
    The complexity arising from these distinct origins can lead to large, difficult-to-understand attack surfaces, presenting a security concern when considering both exfiltration and data schizophrenia.
    Grammar inference can aid in describing the practical language generator behind examples of a data format. However, most grammar inference research focuses on natural language, not data formats, and fails to support crucial features such as type recursion.
    We propose a novel set of mechanisms for grammar inference, RL-GRIT\footnote{RL-GRIT may be pronounced as ``Real Grit.''}, and apply them to understanding de facto data formats. After reviewing existing grammar inference solutions, it was determined that a new, more flexible scaffold could be found in Reinforcement Learning (RL). Within this work, we lay out the many algorithmic changes required to adapt RL from its traditional, sequential-time environment to the highly interdependent environment of parsing. The result is an algorithm which can demonstrably learn recursive control structures in simple data formats, and can extract meaningful structure from fragments of the PDF format. Whereas prior work in grammar inference focused on either regular languages or constituency parsing, we show that RL can be used to surpass the expressiveness of both classes, and offers a clear path to learning context-sensitive languages. The proposed algorithm can serve as a building block for understanding the ecosystems of de facto data formats.

    
    
    
    
\end{abstract}


\section{Introduction}\label{sec:intro}



For data formats of practical complexity, learning a correct grammar from data alone is an impossible problem due to missing semantic information.
Fortunately, oftentimes a desire to learn a grammar from data comes not from complete ignorance of the format, but from a desire to better understand specific usages of the format.
With large, well-adopted formats, such as {\em Portable Document Format} (PDF), an ecosystem forms with programs for reading and writing files in the format. The programs in this ecosystem do not always adhere to the specification's original intent, and sometimes add features or contain bugs, resulting in a modified, de facto specification which is difficult to understand. Understanding these modifications requires grammar inference algorithms which can represent and capture the complicated data structures often found in data formats. 

Even fragments of the nebulously defined de facto grammar might be helpful. It is common to try to secure enterprise systems by pre-processing data before passing it to an application. In these cases, since the pre-processing code is separate from the application, it is very difficult to ensure compatibility. Analyzing and understanding a partial grammar constructed against real-world examples helps with defining what the application accepts. A format designer might also utilize data-derived grammar information to help design a more robust version of their format, without invalidating pre-existing files. A security researcher might use the same information to aid in understanding how format data flows through a specific parsing program in order to better understand potential vectors of attack. Consider a case in which two programs, such as an operating system's built-in PDF viewer and Adobe reader, render a file two different ways. By analyzing a grammar specifically derived from files which produce this parser differential, and comparing it to a grammar from files which render the same, the security researcher can gain a better idea about which parts of the programs and format are involved. There is also the potential for partial grammars to be leveraged for fuzzing purposes, wherein the fuzzer would look at parts of the grammar which allow for alternation and substitute surprising values instead of the expected values. In all of these cases, being able to learn even fragments of a grammar adds to the pool of information available.

\begin{figure}[b!]
    \centering
    \includegraphics{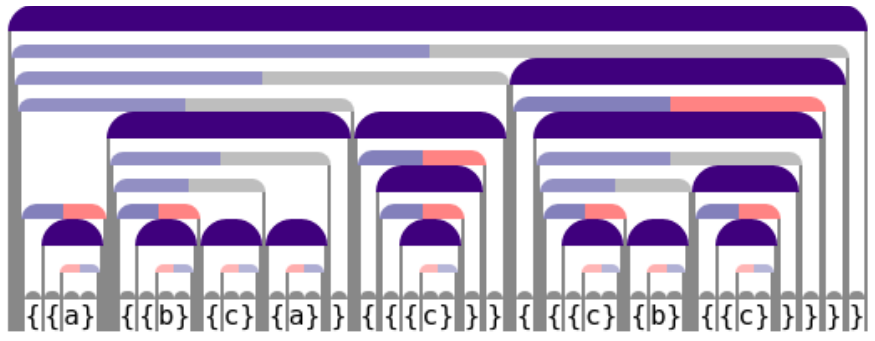}
    \caption{A parse from one of the learned parsers. Here, a recursive type structure was found within a simple JSON-like data format. Different colored bars above the sentence atoms denote different types of actions, as described in \cref{sec:proposed:actions}. Notably, a solid blue bar is a merge (concatenation) of two atoms, a blue bar with a red half is a merge which replaces the red part with a subgrammar atom (defined later in this paper), and a blue bar with a gray half is an anchored merge which omits the gray constituent. This parse is discussed further in \cref{sec:results:simplejson}.}
    \label{fig:intro:exampleparse}
\end{figure}

\begin{figure*}[t!]
    \centering
    \includegraphics{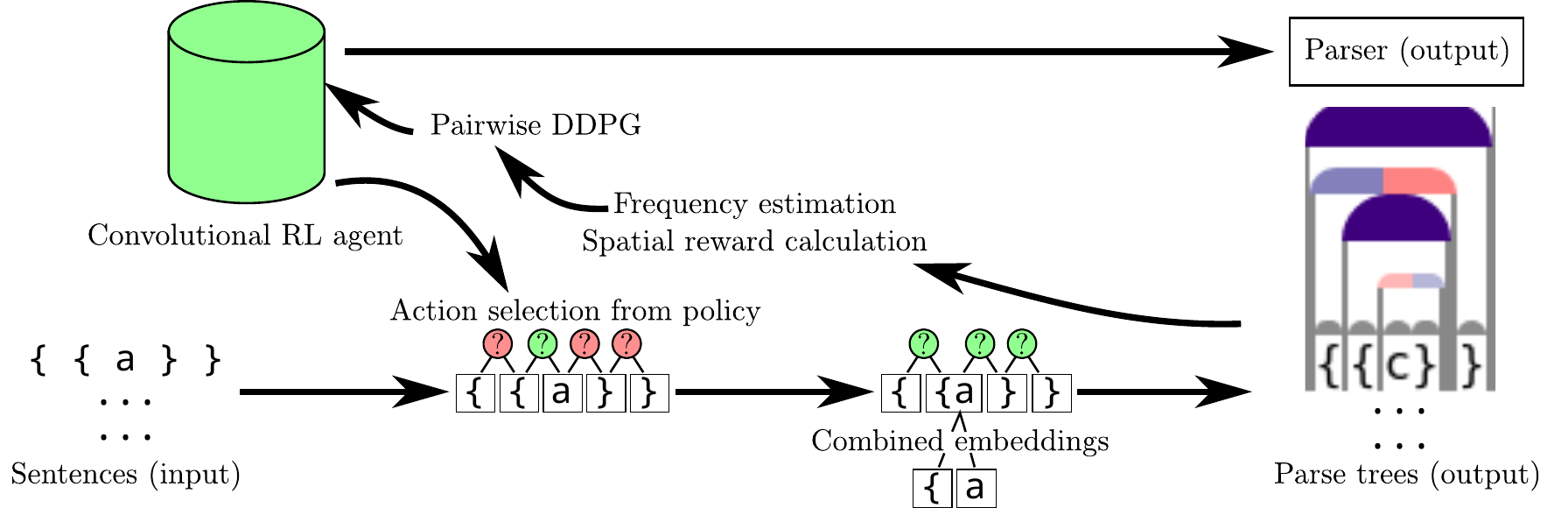}
    \caption{Overview of the RL + grammar inference approach. The proposed method required innovation in the following areas: adapting an RL agent to a convolutional environment; designing actions suitable for learning a parser on arbitrary data; generating deterministic embeddings as atoms are composed through parsing actions; estimating the frequency of atoms within the training data, without being overly dependent on the current policy; adapting RL from the temporal valuation to spatial valuation; and a new loss function, the pairwise DDPG, to accelerate convergence.}
    \label{fig:intro:overview}
\end{figure*}

Unfortunately, existing grammar inference algorithms are very limited in scope, often focusing either on regular languages \cite{carrasco1994rpni} or constituency parsing \cite{drozdov2019diora,drozdov2020sdiora,wang2019treetransformer}. In the following, we use the term ``sentence'' to refer to a string of characters being parsed, and ``atom'' as any singular element in the string being parsed. This is somewhat different from a ``token'' passed to the grammar inference algorithm, in that an atom may be a symbol resulting from previous parsing steps. Algorithms which learn regular languages work well when the training sentences are short and characteristic, but struggle as sentences grow in size \cite{carrasco1994rpni}. Furthermore, regular languages are not as expressive as context free languages, and cannot be used to describe data type recurrency often found in data formats, such as nested dictionaries. At the other end of the grammar inference spectrum, constituency parsing focuses only on merging adjacent, related atoms within a sentence to form larger atoms \cite{wang2019treetransformer}. This works well for small sequences inside of sentences of arbitrary length, but is designed primarily for {\em Natural Language Processing} (NLP). It does not have the expressiveness for data type recurrency, as atom sizes can only grow throughout the parse.

Instead, we investigate {\em Reinforcement Learning} (RL) as the basis for a novel, flexible grammar inference scaffolding. RL is a wonderful {\em Machine Learning} (ML) paradigm for searching spaces of policies. That is, given a sequence of environmental observations, RL seeks to find the best action sequence to perform in service of optimizing some reward. Typically, this is applied to situations such as a robot navigating a maze based on sensory inputs. However, RL provides a generic framework for efficiently enumerating policies consisting of discrete actions, making it one of the most flexible paradigms for searching any decision making process. Success on arbitrarily large search spaces have been achieved when {\em Neural Networks} (NNs) are used to track expected rewards \cite{mnih2013atari}. A series of grammar-relevant actions were identified, and parsers such as that demonstrated in \cref{fig:intro:exampleparse} were able to be learned automatically, solely from data. Applying RL to the grammar inference problem was initially presented at LangSec 2020 \cite{cowger2020icarus}, and the idea has since been significantly expanded and refined in the present work.

Grammar inference is a difficult problem, and the purpose of this work was to explore if that complexity could be represented within RL as a series of actions over atoms in a sentence. This is similar to how existing parsers work, converting a flat sentence into a richer tree (or graph) representation, also known as an {\em Abstract Syntax Tree} (AST). Unlike most existing parsers, which track only complete data structures, grammar inference requires a notion of partial progress, which is achieved in this work via a bottom-up (as in constituency) parsing algorithm. Inference of a data format must also learn to separate control bytes, which denote the format's layout, from data bytes, including user data which may contain any subgrammar, including NLP. Categorizing arbitrary bytes within a file as relevant to the format, as opposed to being relevant only within a given string, is a difficult task.

Designing an algorithm to learn to identify fragments of structure within a grammar using RL required various innovations in both the RL and grammar inference spaces, including: a convolutional RL agent, for location-invariant parsing of atoms given arbitrary parsing actions; a set of data format-appropriate actions from which a grammar might be composed; a deterministic means of generating higher-order embeddings for compound atoms during the parse; an algorithm for estimating the frequency of compound atoms which must be counted alongside learning the policy; a new reward-propagation mechanism for RL based on spatial relations instead of temporal relations; and a new loss-function for RL with purpose-built convergence properties. The overview of the whole approach, including these innovations, is illustrated in \cref{fig:intro:overview}. These improvements work in concert to demonstrate how parsers might be learned via RL.

To demonstrate the algorithm's potential for supporting format experts, results on two toy formats and an excerpt of the PDF format are presented. Since we are pursuing qualitative insights, we do not focus on many metrics in this particular work. Rather, what is provided is a proof-of-concept for a new family of grammar inference algorithms based on RL, which possess flexibility far surpassing state-of-the-art grammar inference algorithms.

\section{Adapting reinforcement learning to grammar inference}

The ideas behind RL are well-documented \cite{mnih2013atari,watkins1992qlearning,mnih2016a2c}, so many of them will be glossed over in this article. We present only the ideas necessary for understanding the proposed modifications to the underlying RL algorithm. This section will focus on a quick primer of RL, followed by the proposed mechanisms by which RL may be adapted for the grammar inference task.

Many of the techniques and results in this work should be considered a preliminary entry in a new family of grammar inference algorithms. What is presented demonstrates that RL may be effectively modified to learn complicated grammars from data, though we imagine many parts of the algorithm could be further improved.

\subsection{Reinforcement learning}\label{sec:proposed:rl}

In RL, an agent interacts with an environment through an imperative loop which consists of sensing the environment, deciding an action based on the sensed environment, and measuring the result as a single scalar, the reward signal. This behavior is looped until the agent either completes the task or is otherwise rendered inoperable. The goal of RL is to discover a policy for action selection which maximizes the sum of rewards across all time steps. It has been shown that for situations where the Markov assumption holds, meaning the result of an action may be characterized entirely by available observations, this may be solved by simulating the environment and tracking the expectation of all possible combinations of state and action via the Bellman equation \cite{watkins1992qlearning}. Once the table of state and action combinations is converged, the optimal policy $\pi^*(x)$ is the one corresponding to actions which maximize the Bellman equation:

\begin{align}
    V(x) &= E\left[ \max_{a \in \Gamma(x)}\left\{F(x, a) + \lambda V(T(x, a))\right\} \right], \label{eq:rl:bellman} \\
    \pi^*(x) &= \argmax_{a \in \Gamma(x)} \left\{F(x, a) + \lambda V(T(x, a))\right\}, \nonumber
\end{align}

\noindent where $E[\cdot]$ is the expectation operator, $x$ is the environment's state; $V(x)$ is the expected reward of executing an optimal policy from state $x$, aggregated across both the immediate and future states; $a$ is the action to be executed; $\Gamma(x)$ is the set of immediately valid actions from the environment's state; $F(x, a)$ is the expected immediate reward from taking action $a$; and $T(x, a)$ is the environmental state which follows $x$ after executing $a$. $\lambda$ is known as the discount factor, and is a constant less than or equal to $1$ which is used to control the stability of convergence in practice.

This method of RL is referred to as {\em Q Learning}, and is known as an ``off-policy'' solution to the RL problem, as the final optimal policy does not depend on the policy used for exploration, provided it is sufficiently stochastic to populate the table to convergence. By contrast, an ``on-policy'' solution to the RL problem may be found in {\em Actor-Critic} (AC) formulations \cite{mnih2016a2c}, where a stochastic policy $\pi_t(x)$ is explicitly learned across time $t$ alongside a table resulting from a modified Bellman equation:

\begin{align}
    a_t &\sim \pi_t(x), \nonumber\\
    V_t(x) &= F(x, a_t) + \lambda V_t(T_t(x, a_t)), \nonumber\\
    V(x) &= E\left[ V_t(x) \right], \label{eq:rl:bellman-ac}\\
    \pi_{t+1}(x) &= \begin{cases}
        \text{increase $P(a_t)$}, &\text{if } V_t(x) > V(x) \\
        \text{decrease $P(a_t)$}, &\text{else}
        \end{cases}, \nonumber
\end{align}

\noindent where $V_t(x)$ is the actual observed reward for state $x$.

Full actor-critic algorithms have additional details \cite{mnih2016a2c}, but the key is that the value table cannot converge until the policy converges, and that actions performing better than the current policy become more likely. Eventually, $\lim_{t\to\inf} \pi_t(x) = \pi^*(x)$, assuming adequate exploration.

In summary, RL is a family of algorithms for interacting with an environment in an effort to learn a policy which optimizes some aggregate reward across all actions taken. For RL to be optimal, the Markov assumption must hold, but crucially, the algorithms can come across interesting policies even where it does not.

\subsection{Learning a parser via RL}\label{sec:proposed:parser}

To adapt RL to grammar inference, we circumvent the grammar part entirely and begin with learning a parser. The parser may include lexical structure, such as re-interpreting bytes in the input string as an integer token. The primary reason for this is that any kind of inference benefits from a search space with progressive improvements in the measured reward. All-or-nothing situations lack the descriptive richness required for modern ML techniques to work. Grammars have developed out of a need to exactly specify sets of accepted strings, implying an all-or-nothing design. Focusing on learning a parser allows the RL algorithm to recognize and exploit frequent data structures in the leaves of a data format, digesting the easy bits of the format before tackling the harder, high-level challenges. Compare this to a grammar, which is typically specified top-down, with one production giving way to more productions which specify the entire range of the language. Notably, this is in line with other work on grammar inference in the NLP domain, which tends to focus on constituency parsing \cite{drozdov2020sdiora,wang2019treetransformer}.

\begin{figure}[t]
    \centering
    \includegraphics{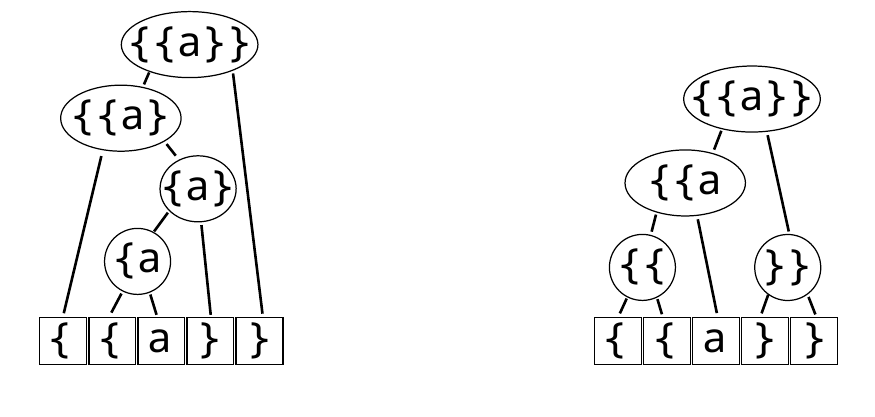}
    \caption{Constituency parsing consists of determining which adjacent atoms have the greatest relation to one another, and combining them into a common parent atom. The process is repeated until a single atom remains.}
    \label{fig:proposed:parser:constituency}
\end{figure}

To learn a parser, we first start with a representation which mirrors the NLP research. In constituency parsing, adjacent tokens are merged together, highlighting common low-level representations or phrases (see \cref{fig:proposed:parser:constituency}). To cast this into the RL paradigm, we might consider parsing a string to be repeatedly merging adjacent atoms into a new atom, until only one atom remains. Therefore, if we have a sentence of $3$ tokens, there are $2$ possible merge actions for the first step of parsing, and only one for the second step. Extrapolating, any sentence of $N$ tokens has $N-1$ possible merges. This problem design is similar to Chomsky Normal Form \cite{chomsky1959normalform}, as it results in a binary tree.

\begin{figure}[t]
    \centering
    \includegraphics{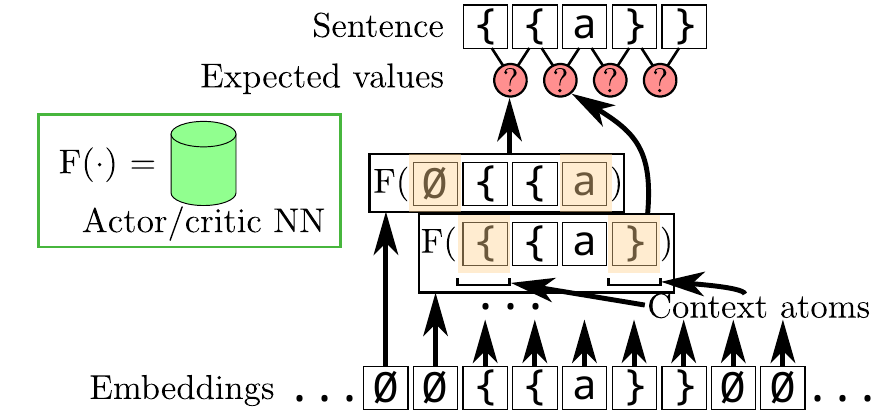}
    \caption{Illustration of convolutional agent evaluation. Each sentence, composed of atoms, has an associated infinite list of embeddings, with embeddings not corresponding to atoms being populated as $\vec{0}$. The agent determines the expected value of different actions by evaluating adjacent overlapping windows of embeddings. Optionally, additional context atoms might be included on either side of the atoms directly affected by an action.}
    \label{fig:proposed:parser:convolution}
\end{figure}

Our RL agent therefore must try to learn a policy based on the current atoms in the string, and must choose at each step which merge should be performed based on that state. A principal concern in the parsing of data formats is that any low-level data structure should be parsed the same regardless of its location in the string. For example, if we consider an excerpt in the JSON format, \mintinline{text}|{"a": [0], "b": [1]}|, both \mintinline{text}{[0]} and \mintinline{text}{[1]} should parse as array types, regardless of their respective key associations and location in the string. In ML, this might be achieved via a convolution \cite{krizhevsky2012imagenet}, where a function is applied to multiple overlapping windows, but function evaluations do not affect one another across different windows. Thus, the value table from the Bellman equation would be a function of a fixed number of adjacent atoms, rather than a function of the entire string. This is illustrated in \cref{fig:proposed:parser:convolution}.

The reward function we use is covered in \cref{sec:proposed:reward}, but it is essentially based on the log frequency of each atom resulting from an RL action; there are additional caveats which will be covered after introducing a few other concepts. Importantly, the convolutional method of evaluating rewards violates the Markov assumption. If we consider two strings in a simple, JSON-like format, \mintinline{text}|{a}| and \mintinline{text}|{{a}}|, then we see that the actions allowed after merging \mintinline{text}|'{'| and \mintinline{text}|'a'| depend on context outside of just those two atoms: in the first example, the resulting atom may only be merged with \mintinline{text}|'{'|, while the second example allows for the subsequent merging of either \mintinline{text}|'{'| or \mintinline{text}|'}'|. However, it is a simplification which allows us to respect other locality rules of data formats, and those locality rules in turn increase the sample efficiency of the ML algorithm. This violation of the Markov property may somewhat be mitigated by parameterizing the value table on additional context atoms, as in \cref{fig:proposed:parser:convolution}. However, for an RL agent to capitalize on the repetition of low-level data structures, convolution is an important facet of applying RL to grammar inference, and too much context could lead to overfitting and poor generalization abilities.

At this point, a basic framework for replicating existing constituency parsing work has been described, using RL instead of the techniques used by state-of-the-art algorithms. To surpass the power of constituency parsers, we must add additional variety in the set of available parsing actions.

\subsection{Data-friendly actions}\label{sec:proposed:actions}

The base RL agent learns to associate each local context of atoms with a merge reward, and can repeatedly take the merge corresponding to the maximum predicted reward to parse the sentence into a binary tree with the maximum reward (caveats in \cref{sec:proposed:reward}). However, one of the motivating aspects for using RL is that it supports arbitrary actions, so long as the reward function can accommodate them. If the action set were limited to merges, it would be difficult to beat the performance offered by state-of-the-art NLP algorithms, as constituency parsers are well studied. However, looking at the output from constituency parsers, as in \cref{fig:proposed:parser:constituency}, it is clear that limiting the action set to merges will always result in an inability to capture high-level structures, as a direct result of the growing size of atoms in the string. For example, consider the following simple, JSON-like format, referred to as Simple-JSON for the rest of this work:

\begin{minted}{text}
S -> '{' ('a' | 'b' | 'c' | S+) '}'
\end{minted}

Like JSON, Simple-JSON supports recursive data structures. The Simple-JSON snippet \mintinline{text}|{{a}}| may be parsed by merging \mintinline{text}|{a}| into a single, commonly occurring atom, but this then results in the type of the inner structure \mintinline{text}|'{a}'|, defined by the representation of its atom, being different from the type of the outer structure \mintinline{text}|'{{a}}'|. In the Simple-JSON specification, both of these are instances of the production rule defined by \mintinline{text}|S|, and the merge action alone is not capable of representing that recurrent relation.

The above might be viewed as a type system over unique atoms discovered throughout the parse. Wherever a novel combination of letters is formed, this is analogous to a new type, and the parser might re-prioritize its decisions based on the new type's presence in the sentence being parsed. Therefore, to allow for recursive data structures, we require a mechanism that results in the same types seen earlier in the parse.

To support higher-level abstractions, we propose and motivate two additional types of actions: the anchored merge, and the subgrammar replacement merge, pictured in \cref{fig:proposed:actions:merges}. Note that these are not expected to be optimal from a language theoretic point of view, but that they do capture aspects which occur in practical data formats, and are a viable starting block for exploiting the flexibility of the RL representation for grammar inference. Notably, the anchored merge is designed to fill the a similar role to the Kleene star, and the subgrammar replacement merge is designed to fill a similar role to the alternation operator.

\begin{figure}[t]
    \includegraphics{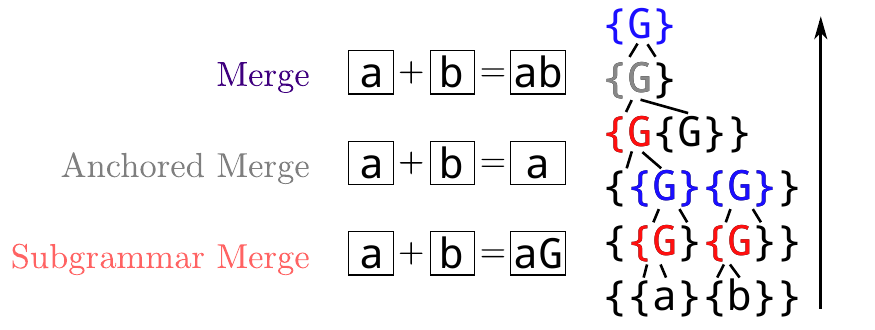}
    \caption{The actions available to the RL-based grammar inference system are the merge, anchored merge, and subgrammar merge, modeled from constituency parsing, the Kleene star, and the alternation operator, respectively. On the right is an example parse of a simple, JSON-like sentence which combines these actions to achieve meaningful data type recurrence. Action color coding matches \cref{fig:intro:exampleparse}.}
    \label{fig:proposed:actions:merges}
\end{figure}

The anchored merge takes two atoms, \mintinline{text}|'A' 'B'|, and merges them into the single atom whose representation is the same as either \mintinline{text}|'A'| or \mintinline{text}|'B'|, depending on whether the anchored merge is left- or right-biased. We will write this as \merge{\mintinline{text}|'A' 'B' -> 'A'|} for a left-biased anchored merge. In the Simple-JSON example \mintinline{text}|{{a}}|, we might consider using an anchored merge to collapse \mintinline{text}|'{' 'a' -> '{'|. If we do that, then we might next perform the merge \mintinline{text}|'{' '}' -> '{}'|, and could then use another anchored merge of \mintinline{text}|'{' '{}' -> '{'| to allow a second occurrence of the \mintinline{text}|'{}'| atom at the top level. Since this top-level atom has the exact same type (representation) as the atom at a lower level, type recursion has been achieved. Anchoring also allows for the learned parser to handle Kleene star semantics, as \mintinline{text}|'a' 'b' -> 'a'| allows an \mintinline{text}|'a'| atom to consume as many neighboring \mintinline{text}|'b'| atoms as desired.

While the anchored merge allows for data type recursion and Kleene star behaviors, they don't allow for the alternation operator. In the Simple-JSON language, each set of curly braces which contains a letter (\mintinline{text}|a|, \mintinline{text}|b|, or \mintinline{text}|c|) contains exactly one letter. With anchored merges we can collapse the type down, but we cannot capture the ``exactly one'' constraint. To address this, we have introduced subgrammar replacement merges, where one of the two atoms being merged has its representation replaced with tokens not occurring in the input language; e.g., \merge{\mintinline{text}|'A' 'B' -> 'AG'|} for a left-biased subgrammar replacement. The Simple-JSON case then becomes \mintinline{text}|'{' 'a' -> '{G'|, which becomes the type atom \mintinline{text}|'{G}'|, and may be combined with anchored merges to both enforce the exactly-one constraint and allow for a recursive data representation.

An ideal Simple-JSON parse using these actions is shown in \cref{fig:proposed:actions:merges}.

\subsection{Embedding generation}\label{sec:proposed:embedding}

In practice, the value table used for RL is typically approximated with an NN which takes the state as input and outputs the information required to propagate future rewards in \cref{eq:rl:bellman} and execute a policy. 
In specifying the NN's input, vector embeddings are a common vehicle for dealing with vocabularies \cite{mikolov2013word2vec}. For example, if we know that our vocabulary is made up of \SI{100}{words}, then \SI{100}{} random vectors are generated in an $N_{emb}$-dimensional space. Often, these embeddings are allowed to drift throughout training to better optimize the objective function.

In the grammar inference case proposed in \cref{sec:proposed:parser}, an issue arises as a result of not being certain of the final vocabulary size. That is, when we merge \mintinline{text}|'A' 'B' -> 'AB'|, the correct embedding for \mintinline{text}|'AB'| is an open question. One approach would be to simply generate a new embedding vector for each new type of atom, arbitrarily growing the vocabulary size. In practice, this might result in a lot of embeddings. Consider the sentence $V_0V_1...V_v$, where $v$ is the sentence length, which also happens to be the vocabulary size. There are a variety of potential atom types in this sentence: as specified in \cref{sec:proposed:actions}, each token may either be retained, replaced with a subgrammar symbol, or ignored, as we merge the sentence down to a single atom. The result is roughly $3^v$ possible atom types, which quickly becomes intractable to store. 

Instead, we propose a deterministic, procedural method for generating the embedding of \mintinline{text}|'AB'| from \mintinline{text}|'A' 'B'| based on arbitrary rotation. This method must A) preserve ordering information of the constituent atom types used to create the new type, and B) result in the same embedding for an atom type regardless of the sequence of actions resulting in that atom type. For example, the same embedding must result from both \mintinline{text}|'A' 'BC' -> 'ABC'| and \mintinline{text}|'AB' 'C' -> 'ABC'|. This may be achieved by deterministically rotating the left atom's embedding, to preserve temporal order, and then summing the rotated embedding with the right atom's embedding.

To achieve this effect, a special orthogonal group is selected in the $N_{emb}$ dimensions. This is functionally an $N_{emb}$-dimensional rotation with guaranteed orthogonality. Optionally, to preserve correlations between input and output embeddings, it may be taken to the $\theta_{emb}$ power, with $\theta_{emb} \le 1$. The resulting rotation matrix is then stored as $\Phi_{emb}$. Each atom type additionally tracks its symbolic length, which for merges is equal to the length of underlying string represented. To combine two embeddings, the left atom's embedding is multiplied by $\Phi_{emb}$ taken to the power of the symbolic length of the right atom, and then the right atom's embedding is added to the result. This result is divided by the square root of the new atom's symbolic length (the sum of its consistuent atoms' symbolic lengths), to keep the expected magnitude consistent for all embeddings regardless of symbolic length, and is the embedding for the resulting atom. For subgrammar merges, the atom being replaced with a subgrammar token is replaced with an atom type formed from a special subgrammar vocabulary element with a fixed symbolic length of 1. For anchored merges, whichever contributing atom is kept gets its embedding information copied to the new atom.

This algorithm allows for biased correlation between compound atoms and their constituent parts, retains order information to ensure the embeddings for \mintinline{text}|'AB'| $\ne$ \mintinline{text}|'BA'|, and will procedurally generate the same embedding for long atom types regardless of the action branches used to create the long type.





\subsection{The value table}\label{sec:proposed:reward}

The value table for our proposed RL-based grammar inference algorithm is approximated via a convolutional NN. Unlike the convolutional NNs used in ML vision literature \cite{krizhevsky2012imagenet}, which use computation blocks that pass data between adjacent features, ours is applied as a sliding window used for input selection. This distinction enforces that each reward be parameterized only by the atoms being considered for one of the merge action types as well as a few additional neighboring atoms for context. The resulting parser runs in $O(N \log N)$ time, as a consequence of expected rewards being cached and kept in a sorted list. 
Using a sliding window scheme implements only part of the RL adaptation required for grammar inference, as several additional facets relating to the value table need to be addressed: the immediate reward signal per action, a means of approximating a non-stationary reward function, and compensation for non-sequential actions.

\subsubsection{Immediate reward signal per action}

For the reward resulting from each action, we consider the average log frequency of the atom resulting from a merge with the same constituent atoms. For example, if we merge \mintinline{text}|'A'| and \mintinline{text}|'B'|, then we would look up the frequency of \mintinline{text}|'AB'| and use its log as the reward to the agent, regardless of the merge's type (that is, the same base reward is used for a standard, anchor, or subgrammar merge). Intuitively, when summed across all atoms in the final parse tree, this log reward is analogous to the negative of the information content contained in the atoms created during the sentence's parse, with the optimal parse being the one resulting in the least information content (as measured by surprising atoms). That is, we want the parse structure to contain as much information as possible, removing that information from the individual atoms formed throughout the parse. 

With the anchored and subgrammar merge actions, an optimal but naive solution is to anchor merge all atoms into the most common atom, maximizing repetition and minimizing surprise. To counteract this naive solution, we additionally weight rewards for these actions by adding the log of $\alpha_{anchor}$ or $\alpha_{subgrammar}$, two scalars less than 1.

\subsubsection{Non-stationary reward}

While the above sounds reasonable, it's actually impossible to unambiguously determine the frequency of each possible atom resulting from a merge. Consider the input string \mintinline{text}|{{a}}|: what is the frequency $F($\mintinline{text}|'{G}'|$)$ when overlapping occurrences cannot be differentiated? It could be 1 if we only end up with e.g., \mintinline{text}|'{' '{a}' -> '{G'|, 2 if we find each matching set of braces, or 4 if we allow for each outer brace to also pair with its opposing inner brace.
That is, the frequency of an atom depends on the parsing policy. However, our task is to search policy space to find the best parser, and a non-stationary definition of frequency based on the current policy would result in wildly different optimal policies, dependent upon the initial policy selected for the search. 

Instead, we consider what the maximum count might have been with any policy in the neighborhood of the current policy. This is accomplished by associating each action type with a list of atoms to be counted, evaluating the probabilities of actions which do ($P(good)$) and do not ($P(bad)$) result in each atom being counted, and logging $(P(bad) + P(good)) / P(good)$ as the number of times an atom was seen for frequency counting purposes. This corrected frequency statistic is exponentially averaged to smooth out peaks, with a decay of $e^{-1}$ every $T_{freq}$ input characters. However, the decay between two subsequent observations of an atom is limited to $e^{-1/N_{freq}}$, to yield more accurate estimates for low-frequency atoms. We use the same exponential averaging to keep track of the uncorrected frequency of observations for each atom type, and use that value as a high water mark for which the associated corrected frequency is the value used. To prevent the probability for atoms resulting from anchor merges approaching 1 on larger datasets, each anchor or subgrammar merge resulted in a $e^{-1/2}$ penalty applied to subsequent atom counts. To avoid over-counting across different combinations of actions which result in the same atom, the ratios of $P(good)$ and $P(bad)$ are propagated from constituent atoms to their parents as several separate statistics, which are used by descendant atoms based on relative positioning. The full process for approximating an atom's frequency is executed each time an action is performed that triggers an observation of that atom, and is detailed in \cref{alg:proposed:reward:frequency}. 
During training, for the first batch any atom is seen, a frequency value of $1/T_{freq}$ is used.

\begin{algorithm*}[t!]
    \SetArgSty{}
    \SetKwInOut{Input}{Input}
    \SetKwInOut{Output}{Output}
    
    \Input{$S$, a sentence of atoms; $A$, an atom resulting from the action being taken; $E$, the list of atoms being used to produce the unique representation of the resultant atom $A$ (notably, including the subgrammar symbol where appropriate); $P(\cdot)$, a function for getting a scalar proportional to the probability of an action given the current policy.}
    \Output{$C_{outer}$, the count of atoms with a representation equal to the concatenation of $E$ to log for frequency counting, as well as the count for descendants which only have one constituent; $C_{left}$, the count for descendants which only add atoms on the right; $C_{right}$, the count for descendants which only add atoms on the left; and $C_{inner}$, the count for descendants regardless of where they add atoms. These are stored on resulting atom as e.g. $A.C_{outer}$.}
    
    \If{$E$ consists of a single atom in $S$}{
        \emph{anchored merges produce an atom which is derived from a single constituent} \;
        \Return{Same $C$ statistics from $E$.}
    }
    
    $C_{outer} \leftarrow 1$, $C_{inner} \leftarrow 1$, $C_{left} \leftarrow 1$, $C_{right} \leftarrow 1$ \;
    \ForEach{Atom $T$ from $S$ participating in $E$}{
        \If{$T$ is left-most in $E$}{
            Update stats s.t. $C_{outer}$ and $C_{left}$ are multiplied by $T.C_{left}$; other statistics multiplied by $T.C_{inner}$.
        }
        \ElseIf{$T$ is right-most in $E$}{
            Update stats s.t. $C_{outer}$ and $C_{right}$ are multiplied by $T.C_{right}$; other statistics multiplied by $T.C_{inner}$.
        }
    }
    
    \emph{Create lists of ``good'' and ``bad'' probabilities which would affect the count of $A$.} \;
    \emph{Note that $G_{all}$ and $B_{all}$ are used to mean: append to all $G$ and $B$ lists, respectively.} \;
    $G_{outer} \leftarrow []$, $G_{left} \leftarrow []$, ..., $B_{outer} \leftarrow []$, ... \;
    \ForEach{Action $Z$ which shares constituents with $E$}{
        \If{$Z$ shares all constituents with $E$}{
            \lIf{the atom resulting from $Z$ matches $A$}{
                AppendTo($G_{all}$, $P(Z)$)
            }
            \lElse{
                AppendTo($B_{all}$, $P(Z)$)
            }
            
        }
        \Else{
            \emph{Only 1 overlapping atom} \;
            \If{This action does not replace the overlapped atom with an equivalent atom}{
                \If{Part of this action's new atom matches the representation of the overlapped atom}{
                    $side \leftarrow$ ``left'' if $Z$ extends to the left of $A$'s constituents, ``right'' if extends to the right \;
                    AppendTo($B_{outer,side}$, $P(Z)$)
                }
                \lElse{AppendTo($B_{all}$, $P(Z)$)}
            }
        }
    }
    
    Multiply each $C$ statistic by $1 + (\sum{B}) / (\sum G)$ for corresponding $B$ and $G$ \;
    \Return{New $C$ statistics}
    
    \caption{Frequency estimation algorithm}
    \label{alg:proposed:reward:frequency}
\end{algorithm*}


While this algorithm provides a very rough estimate of the frequency of atoms encountered during the parse, it works well enough in practice for the initial results presented in this work.

Now that the immediate reward is defined, we must define the overall policy reward, which differs from traditional RL due to the non-sequential model of actions used within our parser.

\subsubsection{Non-sequential actions}\label{sec:proposed:reward:actions}

Standard RL can leverage \cref{eq:rl:bellman} to make optimality guarantees in certain conditions.
One requirement for the optimality guarantee is that \cref{eq:rl:bellman} accounts for all delayed rewards resulting from immediate actions via the future term attached to $\lambda$. In particular, when computing the value of an action, the impact of some future action's reward is scaled by an exponentially-decreasing function of the time that passed between the two actions.
However, the proposed, RL-based grammar inference has an unclear relationship between the time at which an action is performed and the related future rewards. Instead, we focus on the distance between actions in the resulting parse tree.

\Cref{fig:proposed:reward:spatial-vs-temporal} demonstrates how the RL-derived parser does not have the same well-defined temporal relations between actions. Simply, a sequence of 3 merges might have the first merge at the beginning of a long sentence, the second merge at the end, and the third merge again at the beginning, absorbing the atom resulting from the first merge. While temporal coherence would place a relationship between the disparate first and second merges, spatial coherence instead allows the algorithm to focus on the relationship between the proximal first and third merges. 

\begin{figure}
    \centering
    \includegraphics{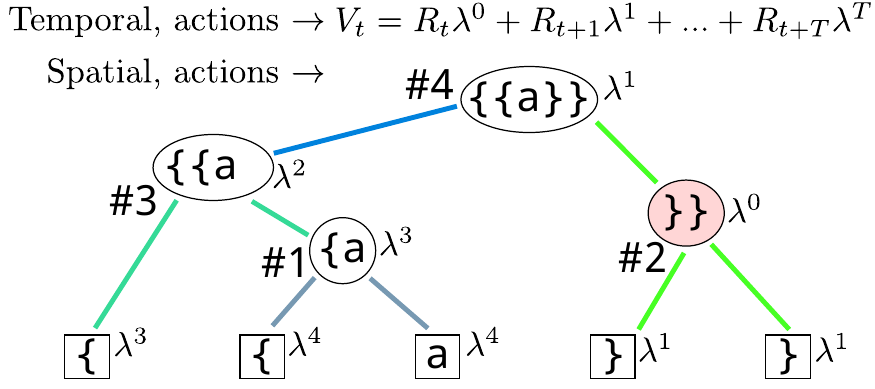}
    \caption{Illustration of the spatially applied discount factor. Whereas standard RL uses a discount factor, $\lambda$, applied temporally to sequential actions, investigating the order of actions, from \#1 to \#4, yields no clear relation during a parse. We instead extrapolate the application of the discount factor to a spatial domain by basing its power on the number of edges between two atoms in the resulting tree. For example, when calculating the value of ``\}\}'', the rewards for ``\{\{a\}\}'' and ``\}'' are multiplied by a discount factor of $\lambda^1$, ``\{\{a'' is multiplied by a discount factor of $\lambda^2$, and ``\{a'' is multiplied by a discount factor of $\lambda^3$. To further control against biases due to the length of the string, we divide the result by the sum of all $\lambda$ terms to get an average reward as the value.}
    \label{fig:proposed:reward:spatial-vs-temporal}
\end{figure}

Thus, rather than a discount factor being applied to the future rewards, as in classic, temporal RL, we apply a discount factor to all constituent atoms, and to the atom of which the new atom becomes a constituent. In other words, the number of edges traversed between two atoms is the power to which the discount factor will be taken when considering the contribution of each node's reward on the other. The result is a scoring system which rewards actions participating in parsers with a high expected reward, without regards to the ordering of those actions. This method works surprisingly well on its own. 



Rather than looking at the sum of all rewards, we use the average, as the number of interactions between the agent and its environment (the string being parsed) depends on the length of the string and not the agent's specific actions. In the Simple-JSON case, this helps normalize against e.g. \mintinline{text}|'}{'| receiving a biased reward simply because it only shows up in longer sentences. Rather than dividing by the length of the sentence, we divide by the sum of all discount factors applied to a node, resulting in a weighted average with the most significant individual contribution coming from the action itself. For example, in a parse tree with 2 actions, the reward computation would look like $R = (1 * I_{action1} + \lambda I_{action2}) / (1 + \lambda)$, where $R$ is the discounted, total reward, and each $I$ is the immediate reward resulting from an action.

\subsection{A new loss function: pairwise DDPG}\label{sec:proposed:dpg}

Classic RL typically uses either Q learning or the actor-critic method. Due to the usage of the average reward instead of the standard weighted sum, the RL-based grammar inference is most amenable to an online algorithm like actor-critic. However, we found the traditional actor-critic model to be too volatile for learning a parser, potentially due to our frequency estimation mechanism. Instead, we designed an approach inspired by the {\em Deep Deterministic Policy Gradient} (DDPG) approach \cite{lillicrap2015ddpg}, which resulted in better convergence for our convolutional agent.

DDPG tracks two quantities for each possible action in each possible context: the likelihood of taking that action under the current policy, stored as a pre-softmax logits value on $(-\inf, \inf)$, and an expected aggregate reward, or value, given a state and the current policy \cite{lillicrap2015ddpg}. On each step, the expected value is updated based on the discounted reward calculated from the current policy, and the policy's distribution is updated such that it maximizes the expected reward. Notably, DDPG differs from actor-critic methods in that poor actions are never explicitly penalized; instead, better policy options are encouraged.

The proposed pairwise DDPG was modified from DDPG for two reasons: to keep the memory and processing footprints small with our convolutional agent, and to encourage the agent to converge away from poor actions quickly.

During a training loop, each time a parsing action is taken, a memory is created with the context embeddings surrounding both the action taken and another action with a higher expected value, where the action with a higher expected value is sampled at random from all such valid actions. That is, we store training memories with just enough information to re-evaluate the chosen action and one probably better action. The actual discounted reward value used for the chosen action is calculated as described in \cref{sec:proposed:reward:actions} after the parse is complete.

During a training loop, we select a random batch of actions from the parsed sentence batch, and use mean-squared-error loss to update the expected value, and use a custom, pairwise loss for the policy update. The policy update algorithm is governed by the loss:

\begin{align}
    P &= (1 + e^{A^* - A})^{-1}, \nonumber\\
    \eta &= \max(0, V^* - V)\frac{P^{\beta}}{\frac{\epsilon}{2} + (1 - \epsilon)P}, \nonumber\\
    L_{policy} &= D(\eta)(A - A^*), \label{eq:proposed:dpg-loss}
\end{align}

\noindent where $A$ is the logit value associated with the chosen action, $A^*$ is the logit value for the action with a higher expected value, $V$ is the expected value of the chosen action, $V^*$ is the expected value of the better action, $\beta$ is a skew constant to bias the algorithm, $\epsilon$ is the probability of the algorithm taking a random action rather than an action selected from the policy, and $D(\cdot)$ is an identity function which detaches its argument from the gradient computation.

Essentially, rather than using a standard softmax-loss across all available actions, we instead sum the loss across pairs of chosen and expected better actions. While typical log-softmax loss results in a pushing magnitude of $1 - P(x)$ on the logit for a selected action $x$, we choose to push based on $1 - P(x|\{x^*, x\})$, and only if $x^*$ is predicted to be a better action than $x$. This change is in contrast to traditional actor-critic loss functions, with which we found negative gradients having a strong effect of reinforcing early biases. The $\beta$th power is involved since we are not interested in storing a probability distribution; the reality of the way in which NNs store the expected reward results in a rotating gallery of best actions. While a traditional log-softmax loss results in the training distribution being learned, we might want something closer to an integrator which goes all-in on the best policy. 

\section{Methodology}\label{sec:methods}

The previous section described an interdependent system of mechanisms to simultaneously estimate the frequency of types within a grammar system and learn a parser which minimizes the information content contained within the atoms generated by that system, capturing as much information as possible in the structure produced by the parser. To demonstrate its efficacy, we applied the system to a few different data formats: Simple-JSON, Simple-JSON inside a stream of meaningless characters, and extracts of PDF dictionaries. Each of these used similar NN architectures and other configuration values.

\subsection{Datasets}\label{sec:methods:datasets}

The Simple-JSON dataset consisted of 128 examples of the grammar, with 8 being withheld from training for evaluation purposes. The generator tracked the depth of each data structure, and had a $0.6^D$ chance of executing the \mintinline{text}|S+| rule for a data structures contents. Here, $D$ is the depth of the current data structure, and was initialized at $1$. If the \mintinline{text}|S+| rule was followed, additional data structures were added in series until $0.6^D$ failed to pass again (encouraging broad hierarchies rather than deep ones). When a letter was chosen as a structure's contents, instead of \mintinline{text}|S+|, then \mintinline{text}|a|, \mintinline{text}|b|, and \mintinline{text}|c| were chosen with equal probability. This dataset was chosen as it has type recurrency similar to that found in real data formats, such as JSON. Random examples from this dataset are: \mintinline[breaklines]{text}|{a}|, \mintinline[breaklines]{text}|{{a}{{b}}}|, \mintinline[breaklines]{text}|{{c}}|, and \mintinline[breaklines]{text}|{{a}{{b}{c}{a}}{{{c}}}{{{c}{b}{{c}}}}}|. 

The Simple-JSON inside a stream dataset, or Simple-JSON-Stream for short, consisted of 512 examples of the grammar, 32 of which were withheld for evaluation. The generator was identical to that used for Simple-JSON; however, each sentence was prefixed with 5-20 characters from the set of all lowercase letters plus \{ and \}. Each sentence was suffixed with another random 5-20 characters from the same set. The purpose of this dataset was to demonstrate that structure could be pulled out of internal bytes in each sentence, without needing structure across the whole sentence. Examples of this dataset are: \mintinline[breaklines]{text}|dtqsc{a}r{hoqozxcuv| and \mintinline[breaklines]{text}|cyarrdkpnts}i}{{{b}}{{c}{b}}}aqhyrrejlt uma qrn|.

The third and final dataset came from $1000$ dictionaries extracted from various PDF files. These varied widely in sentence length: in total, there were \SI{128380}{} characters across the dataset. Fifty dictionaries were withheld for evaluation. Examples from this dataset are: \mintinline[breaklines]{text}|<<\n/Type /Page\n/Parent 447 0 R\n/Resources 456 0 R\n/Contents 455 0 R\n>>| and \mintinline[breaklines]{text}|<<\n/Filter[/FlateDecode]\n/Length1 1426\n/Length2 8801\n/Length3 533\n/Length 9678\n>>|.

\subsection{NN model and other configuration}

To handle a common use case, we also added a ``parse integer'' action to the available feature set, which would replace any number of sequential atoms consisting solely of digits with a single, new integer token. For simplicity, we set the immediate reward for this action to the log of half the frequency of all such integers detected in the training data.

For the model architecture, two residual-style dense networks were used in all experiments, based on the pre-activation ResNet architecture \cite{he2016identity}. One network stored the critic value, and the other the actor value - this separation was to avoid needing to weight the respective losses against one another. Rather than explicit convolutional blocks, dense linear connections were used. All residual block outputs were initialized to zero, as per \cite{he2019tricks}. Each network ended in a dueling head, as per \cite{wang2016dueling}, which aids in separating state information from specific action information.

Unless otherwise mentioned, embeddings were of width 64, and each network layer was twice the embedding width, or 128. Context atoms were not used unless otherwise specified, meaning each network (one for critic and one for actor) took 128 inputs, went through several 128-width layers, one 256-width layer before each dueling head, and finally output 1 number from the state head and 6 numbers from the action head (one for each action type from \cref{sec:proposed:actions} plus one for the integer parsing action). Networks were initialized with Kaiming uniform initialization \cite{he2015delving}.

When context atoms were used, all input weights corresponding to the context atoms were initialized to zero, leaving it up to the learning algorithm to infer relevance. The Kaiming initialization on this initial layer was correspondingly scaled to match the parameter scale without context atoms.

Batch normalization was used as in pre-activation networks. Training occurred on minibatches of 64 memories at a time. The AdaBelief optimizer \cite{zhuang2020adabelief} was used with a learning rate of $5e-4$ for the critic and $5e-5$ for the policy. This differential in learning rate matched the ratio used by DDPG \cite{lillicrap2015ddpg}, and it was found to still be effective with the combination of frequency estimation and the new pairwise-DDPG loss function. We used a step-down learning rate schedule, halving the learning rate 2 times evenly spaced throughout training, until the final minimum learning rate was $2^{-2}$ that of the initial learning rate. All trials were run for 200 epochs over their respective datasets.

During parsing, regardless of training or evaluation, the networks were set in evaluation mode, meaning that each batch normalization layer's statistics were used. Batch normalization was only used during the training step for each minibatch of 64 memories.

During training, each dataset would parse $32$, $16$, and $4$ sentences at a time, respective to each of the datasets from \cref{sec:methods:datasets}, adding the resulting memories into a FIFO buffer of length $10000$. The training mini batches of size $64$ were sampled from this buffer, and the number of batches sampled after each set of training sentences was equal to $\lceil 2N_{\text{memories added}} / 64\rceil$. In other words, on average, each action taken was sampled twice. The buffer was used to lend stability to the policy learning, and was found to be helpful. Frequency estimations were decayed exponentially with $T_{freq}$ of \SI{10000}{}, \SI{10000}{}, and \SI{50000}{} characters for each of the datasets in \cref{sec:methods:datasets}, respectively. The values for $\alpha_{anchor}$ and $\alpha_{subgrammar}$ were set to $0.4$ and $0.5$ for the Simple-JSON-based datasets, and $0.008$ and $0.01$ for the PDF dictionaries dataset. $\beta$ from \cref{eq:proposed:dpg-loss} was set to $0.5$.

\section{Results and discussion}\label{sec:results}

To demonstrate the efficacy of the proposed RL-based grammar inference algorithm, we present results on several datasets. Note that some metaparameters, most specifically the reward function variables $\alpha_{anchor}$ and $\alpha_{subgrammar}$, had a large impact on the results. This follows from the idea that different data formats seek to be efficient at representing different ideas.


\subsection{Simple-JSON}\label{sec:results:simplejson}

The Simple-JSON parsers learned tend to produce parses like the one shown in \cref{fig:intro:exampleparse}. Notably, the inference engine successfully learned the \mintinline{text}|'{G}'| building block of the grammar, and was able to detect its recursive application on example strings. To our knowledge, this result has not been achieved by other research groups, due to state-of-the-art methods being focused more on regular or natural languages rather than the full realm of context-free grammars (further discussed in \cref{sec:related}).

We note that the data does not imbue a bias on the learned parsers --- sometimes they are left-biased, and other times right-biased. This may be seen as an extension of the fact that there is no singular grammar which accepts a given set of strings. Rather, there is always a family of grammars which are varied configurations of the same acceptance machine.

It may also be seen how the type system provided by the actions available to the parser requires visible control atoms, such as \mintinline{text}|'{'|, to anchor content atoms, such as \mintinline{text}|'a'|. That is, as a cost of producing explainable grammars, where each production is tied to precise atoms in the input sentence, the RL + grammar inference system as presented here can only express languages with a visual demarcation of different regions; this is reminiscent of visibly pushdown languages \cite{alur2004vpl}. Most data formats use some tagging mechanism to indicate the type of a union data structure, so we find this to be acceptable, but it is a potential limitation of the approach as presented in this work.

This result demonstrates that the grammar inference algorithm proposed in this work is capable of exploiting recursive structures in data. Next, we demonstrate that these recursive structures might be pulled out even when they are fragments within a larger data format.

\subsection{Simple-JSON in stream}\label{fig:results:json-in-stream}

\begin{figure*}[t!]
    \centering
    \begin{subfigure}{.49\linewidth}
        \includegraphics{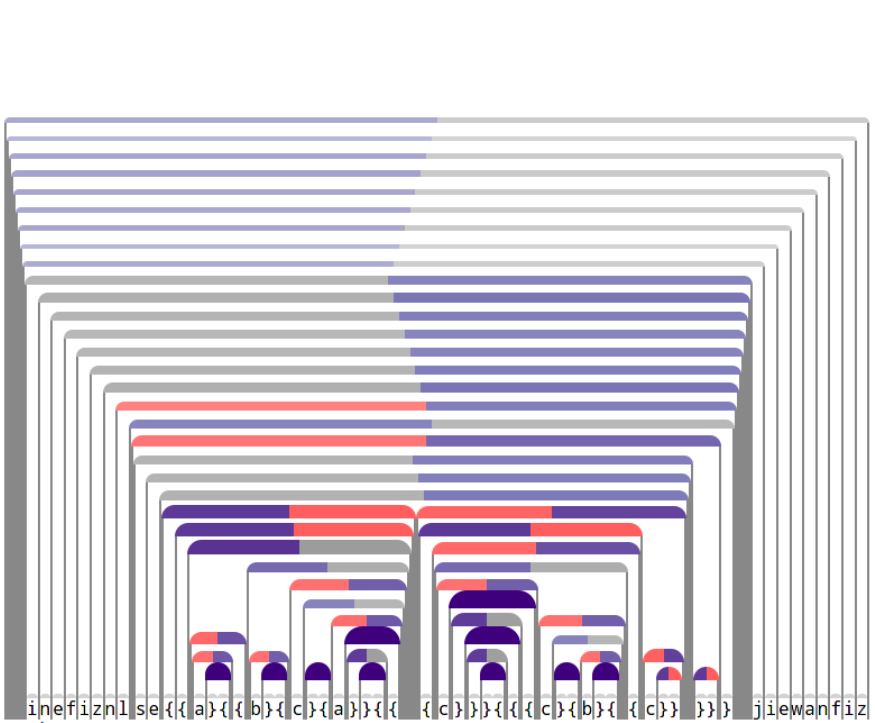}
        \caption{}
        \label{fig:results:json-in-stream:parse:a}
    \end{subfigure}
    \begin{subfigure}{.49\linewidth}
        \includegraphics{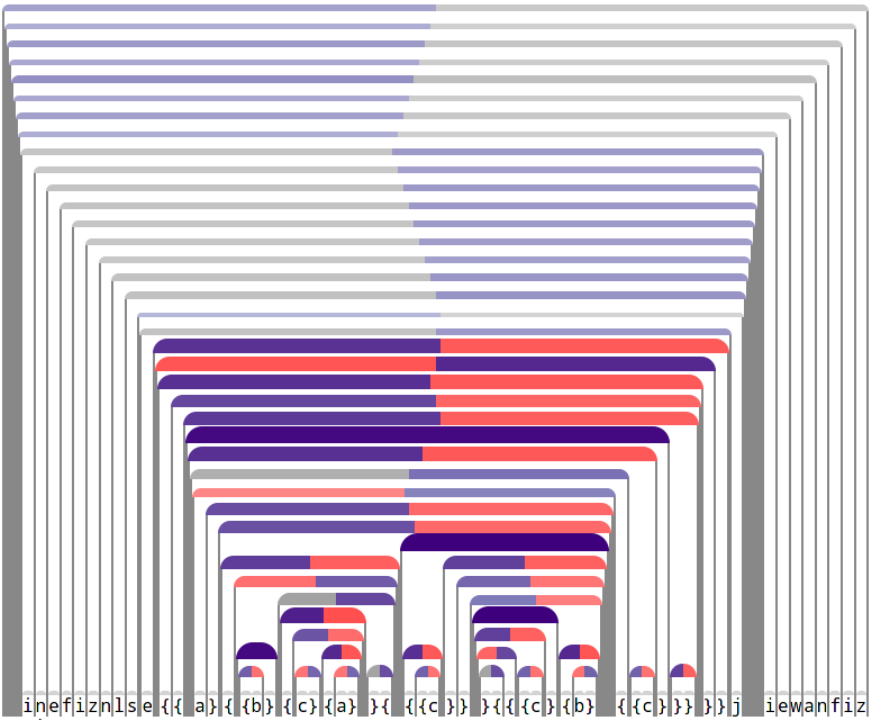}
        \caption{}
        \label{fig:results:json-in-stream:parse:b}
    \end{subfigure}
    \caption{Example parses from parsers learned on the Simple-JSON-Stream dataset, where (a) was trained with no additional context atoms, and (b) was trained with 2 additional context atoms on each side of each action. In (a), the merge to create \texttt{\textquotesingle\}\{\textquotesingle} scores highly, but the algorithm cannot learn the action ordering required to first merge the basic \mintinline{text}|'{G}'| block. In (b), the additional context atoms allow a partial ordering of merge actions to be learned, and thereby some \mintinline{text}|'{G}'| atoms are created during the parse. However, the additional context atoms also resulted in the discovery of patterns which score higher with the chosen reward function, despite leveraging less type recursion.}
    \label{fig:results:json-in-stream:parse}
\end{figure*}

Unlike Simple-JSON, which only has a vocabulary of 3 symbols, Simple-JSON-Stream has a vocabulary of 28 symbols. Additionally, as many of the Simple-JSON data structures are 5 characters or shorter, there is a significant amount of noise relative to the signal of the recursive data structures. To make matters worse, portions of valid data structures, such as \mintinline{text}|'{a'|, often show up in the noisy part of each sentence.

\Cref{fig:results:json-in-stream:parse} shows two example parses trained on this dataset. \Cref{fig:results:json-in-stream:parse:a} used no additional context atoms, while \cref{fig:results:json-in-stream:parse:b} used two additional context atoms on each side of a potential merge. Notably, the spatial reward function from \cref{fig:proposed:reward:spatial-vs-temporal} enforces no ordering over action application. Instead, all actions participating in a well-structured parse get approximately the same reward.
For example, merging \mintinline{text}|'}'| and \mintinline{text}|'{'| is a good action, as their concatenation creates a common atom, but only results in an optimal policy if it follows a subgrammar merge of, e.g., \mintinline{text}|'a'| and \mintinline{text}|'}'|. Otherwise, these brackets get merged before the base block of the Simple-JSON grammar, leading to the bad parse shown in \cref{fig:results:json-in-stream:parse:a}. By adding context atoms, as in \cref{fig:results:json-in-stream:parse:b}, a partial ordering was learned, allowing for the realization of \mintinline{text}|'{G}'| atoms being merged before any potential \mintinline{text}|'}{'| atoms. However, the additional context atoms also allowed the parser to learn fancier behavior, exploiting the reward function to come up with a parse which scored higher than parses leveraging more type recursion (such as the one shown in \cref{fig:intro:exampleparse}). Notably, there were some \mintinline{text}|'{G}'| blocks within this parse.

\subsection{PDF dictionaries}\label{fig:results:pdf}

\begin{figure*}[t!]
    \centering
    \includegraphics{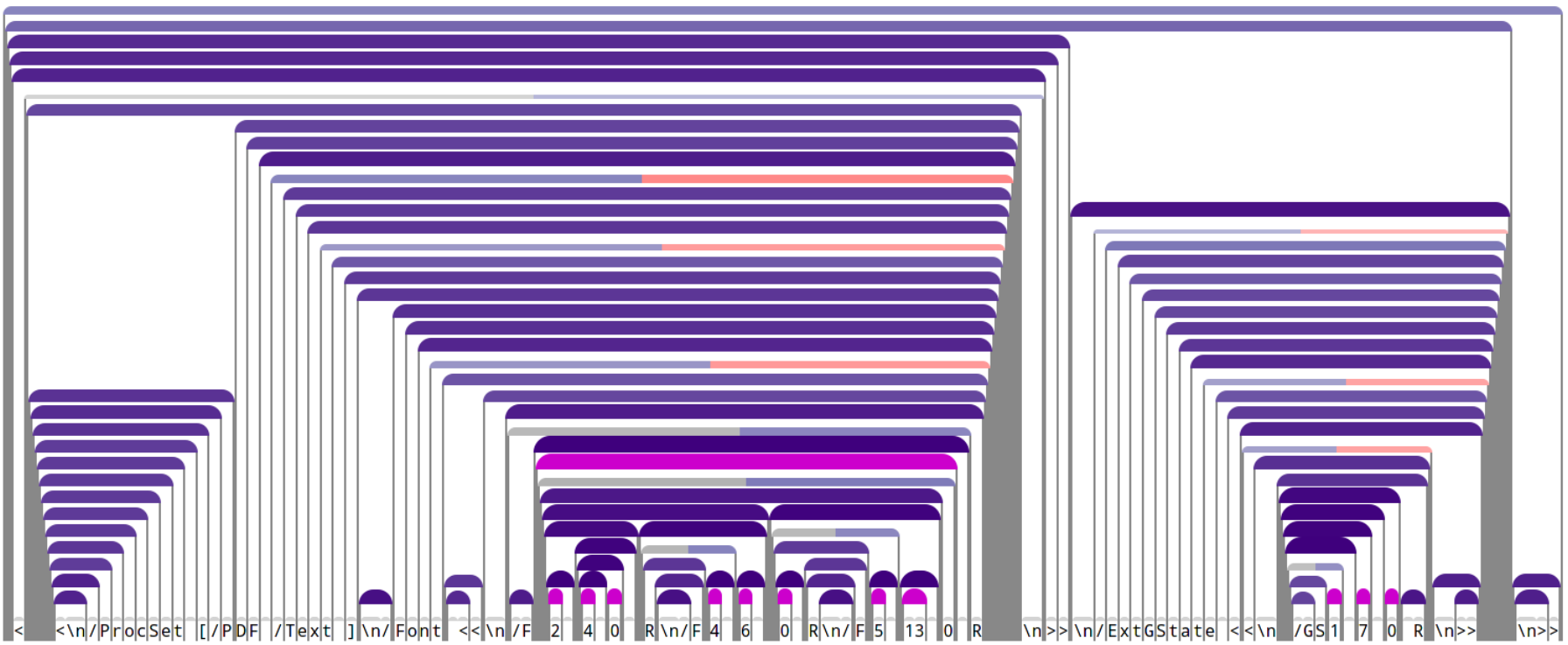}
    \caption{Example parse from a parser learned on the PDF dictionaries dataset. The fuchsia lines indicate the integer parsing action was selected. The learned parser demonstrates having learned several structures, including \mintinline{text}|'/ExtGState~G~'| on the right side (using \mintinline{text}|'~G~'| as a subgrammar symbol), in this case with the subgrammar being an instance of \texttt{\textquotesingle\ <<\textasciitilde G\textasciitilde\textbackslash n>>\textquotesingle}, which is the syntax for a PDF dictionary which includes a final newline.}
    \label{fig:results:pdf:parse}
\end{figure*}

\begin{figure}[t!]
    \centering
    \includegraphics{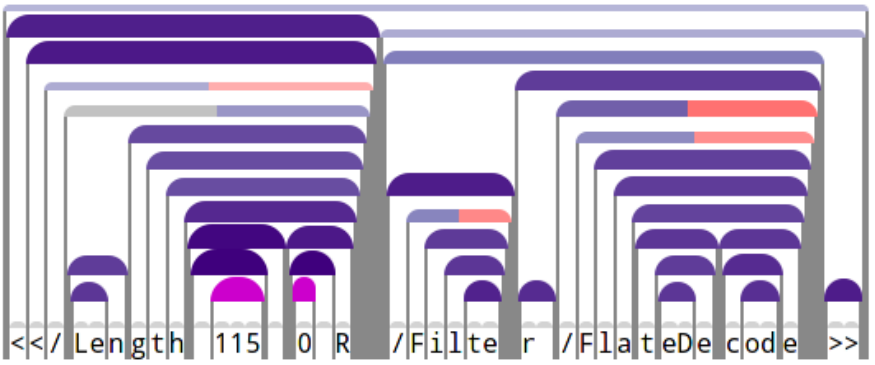}
    \caption{Another example parse from a parser learned on the PDF dictionaries dataset. Object references in the PDF data format have the structure \mintinline{text}|'~INT~ ~INT~ R'|, which was correctly picked up in association with the \mintinline{text}|/Length| key. However, the \mintinline{text}|/Length| key itself was cut in two; it is unclear if this was the result of insufficient training or the reward function used.}
    \label{fig:results:pdf:parse-2}
\end{figure}

\begin{figure*}[t!]
    \centering
    \includegraphics{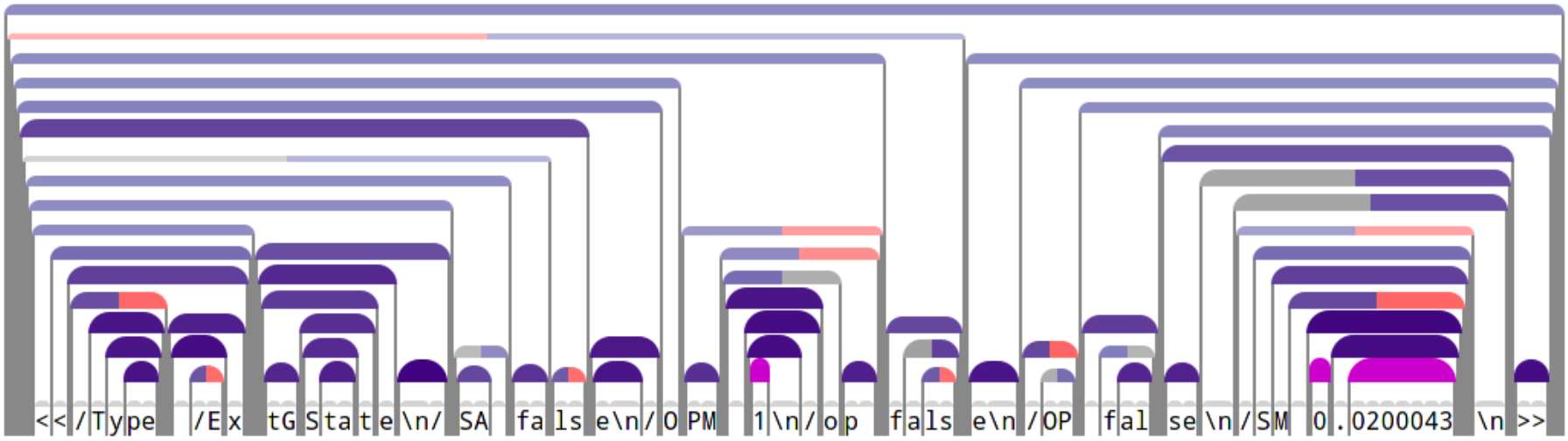}
    \caption{A final example parse from a parser learned on the PDF dictionaries dataset. We include this to show that it learned the atom \mintinline{text}|~INT~.~INT~| for a floating point number, but did not pick up much meaningful structure otherwise. Note that \mintinline{text}|/ExtGState| was not correctly parsed here, where it was in \cref{fig:results:pdf:parse}, due to the different context atoms.}
    \label{fig:results:pdf:parse-3}
\end{figure*}

Finally, we present example parses from a parser learned on a real-world file subformat: dictionaries within the PDF specification. This parser was trained using 2 additional context atoms on each side of each action. 

In the PDF language, dictionaries are always bracketed by \mintinline{text}|'<<'| and \mintinline{text}|'>>'|, and may be nested. In \cref{fig:results:pdf:parse}, it may be seen that while the parser missed the top-level dictionary, it did parse the dictionary following \mintinline{text}|/ExtGState| on the right side as a single atom, \mintinline{text}|'<<~G~\n>>'|. On the left side, atoms like \mintinline{text}|'<\n/ProcSet [/PDF /~G~'| were common in the training data. 

\Cref{fig:results:pdf:parse-2,fig:results:pdf:parse-3} demonstrate additional parses from the same parser as \cref{fig:results:pdf:parse}. Notably, \cref{fig:results:pdf:parse-2} shows the correct parsing of a \mintinline{text}|'~INT~ ~INT~ R'| atom, which corresponds to an object reference in the PDF format. \Cref{fig:results:pdf:parse-3} demonstrates that while some small-scale structures were identified, the parses tend to contain a fair amount of noise, and so additional convergence research is required on the RL-based method of grammar inference.

While we hope to improve this result in the future, we are encouraged by the algorithm uncovering true data structures within the PDF specification.

\section{Related Work}\label{sec:related}

There have been many previous efforts to infer grammars from data. Largely, these are limited to either regular languages or constituency parsing. The proposed method permits a more flexible action set, allowing for more expressive grammars to be inferred than from previous methods. Notably, it provides a window into how context-sensitive languages might be inferred from data through the addition of context-sensitive actions, something not yet achieved by other algorithms. Below, the proposed method is compared and contrasted with several well-known and state-of-the-art methods.

For regular languages, algorithms such as {\em Regular Positive and Negative Inference} (RPNI) can produce finite automata which accept a set of positive training examples and reject a set of negative examples \cite{carrasco1994rpni}. However, the generalizing abilities of RPNI are poor when the training set is not characteristic of the entire grammar, and RPNI cannot make use of statistical information \cite{higuera2010grammatical}. The method we have proposed may also be used to infer regular grammars, but does not require negative examples at this time. However, the proposed method results in a grammar which exploits the statistics of the training set. By contrast, RPNI exploits all characteristics present in the training set, regardless of their frequency. This may be desirable on simple grammars with a complete sample set, but does not scale well to larger grammars.

A notable project focused on inferring data grammars is the LearnPADS project \cite{fisher2008learnpads}. Focused on log data and other record-style formats, the grammar inference part of LearnPADS used minimum description length as a means of recovering semantics from data. The approach was reasonably successful, but could not handle recursive structures. Our approach of minimizing information content in atoms formed during a parse is similar to minimum description length, but more amenable to computing within an RL environment.

Most other work focuses on constituency parsing \cite{wang2019treetransformer,drozdov2019diora,drozdov2020sdiora,kim2019urnng}. These methods are also bottom-up, i.e., they can extract internal data structures without needing to understand the full context of the sentence. To the best of our knowledge, S-DIORA \cite{drozdov2020sdiora} qualifies as the current state-of-the-art with respect to matching constituency trees hand-crafted by language experts on NLP datasets. The DIORA family of algorithms work by building an autoencoder-style prediction task, where each atom is used to predict surrounding atoms. Notably, the implementation of S-DIORA does functionally allow for anchor-style merging, as embeddings for new atoms are generated as a linear combination between the two constituents. However, there is no mechanism for specifying a distinct reward for this type of action, or even identifying when that action is used. Furthermore, S-DIORA cannot always differentiate between atoms such as \mintinline{text}|'AB'| and \mintinline{text}|'BA'| due to this linear combination, something which the proposed method handles with the deterministic embedding combinations from \cref{sec:proposed:embedding}. S-DIORA also provides no mechanism for expansion of the set of available parsing actions. 



\section{Future work}

While the action types presented in this work (merges, anchored merges, and subgrammar merges) can express a superset of the concatenation, Kleene star, and alternation operators, it is likely that they are overly permissive. A more thorough investigation of adapting formal language constructs to the RL approach laid out in this work would be beneficial for leveraging theory work used by designers of data formats.

The parsers learned from this process, when paired with their training data, implement a well-defined grammar. However, it is not human-readable. Subsequent efforts will be required to convert the grammar implemented by the learned parser into a human-readable grammar format.

Supporting context-sensitive grammars is primarily an exercise in designing both appropriate actions for the RL agent, and a mechanism for rewarding the actions appropriately. In a data format, such an action might be to interpret an integer as a length field, and to then consume that many following data structures (atoms). After that, any remaining difficulty comes from the search space being significantly expanded. One option would be to bias the policy space such that this action is only attempted where it makes sense, based on some knowledge of the format under inspection. 

Notably, the techniques presented are amenable to leveraging prior knowledge in a variety of forms. For example, if one has access to a related specification and is looking to adapt it to real-world data, that information might be encoded in the reward function or the agent's initial reward table. If a related specification uses single-character delimiters to surround certain types of data, then parsing actions which result in atoms beginning and ending with the same character might be additionally rewarded to force convergence in that direction. Or, to encourage the network to better explore policies which result in atoms beginning and ending with the same character, without affecting the optimal solution, the value table might be initially biased by adding a fixed constant to the policy network's output for actions which would result in these atoms. That would result in better exploration of the related concept, but the network could eventually overcome this bias.

Work on understanding how data flows through parser executables, such as taint tracking \cite{harmon2020toward}, might be integrated into the reward function to better guide the reinforcement learning process. Taint tracking follows bytes in an input file through the memory of a parsing program, providing a unique window into how data flows between data structures used to process file data at a semantic level. As this information is unavailable when looking only at the file's bytes, we view it as a potentially useful way to align the learned grammar with the actual semantic layout of data files.

\section{Conclusions}

The inference of grammars from data is a well-studied problem in the NLP space, but less so in the realm of data formats. However, even partial progress on data formats can help format experts and security researchers better understand how those formats are used in practice. This work introduced a set of techniques for adapting reinforcement learning to the task of grammar inference, specifically considering the inference of data formats. The initial results are promising enough to merit further study of this approach; the methods presented allowed for the inferring of a recursive type system, which had been out of reach of prior algorithms. A convolutional RL agent was presented for estimating reward values of different parsing actions. A set of data format-appropriate actions were introduced, with motivation coming from the Kleene star and alternation operators. A means of progressively defining embeddings for compound atoms was introduced. A method of propagating RL rewards across space rather than time was introduced to deal with the unique space of abstract syntax trees. Finally, a novel loss function based on DDPG, called pairwise DDPG, was introduced to accelerate convergence of the RL agent's policy. These mechanisms combined form a viable platform for learning parsers from data, built from arbitrary sets of actions. Learned parses were presented on sentences from the Simple-JSON, Simple-JSON in a stream, and dictionaries extracted from PDF files datasets. Future work was discussed, including the support of context-sensitive grammars and integration with taint tracking tools. We believe that RL-GRIT demonstrates an effective initial effort for learning parsers requiring recursive types from only examples of a data format, and hope it might be used to help data format designers better understand de facto usages of their data formats.

\section*{Acknowledgements}

This material is based upon work supported by the Defense Advanced Research Projects Agency (DARPA) under Contract No. HR0011-19-C-0073. Any opinions, findings and conclusions or recommendations expressed in this material are those of the author(s) and do not necessarily reflect the views of the Defense Advanced Research Projects Agency (DARPA).

The author thanks Alexander Grushin and David Darais for their feedback and advice in the editing of this work.


\bibliographystyle{./sty/ieee/IEEEtran-nomonth}
\bibliography{./sty/ieee/IEEEabrv,main}

\end{document}